\documentclass[10.5pt,compsoc]{tst}
\usepackage{graphicx}
\usepackage{footmisc}
\usepackage{subfigure}
\usepackage{url}
\usepackage{multirow}
\usepackage[noadjust]{cite}
\usepackage{amsmath,amsthm}
\usepackage[english]{babel}
\usepackage{amsthm}
\usepackage{amssymb,amsfonts}
\usepackage{booktabs}
\usepackage{color}
\usepackage{ccaption}
\usepackage{booktabs}
\usepackage{float}
\usepackage{fancyhdr}
\usepackage{caption}
\usepackage{xcolor,stfloats}
\usepackage{comment}
\graphicspath{{figures/}}
\usepackage{cuted}  
\usepackage{captionhack}
\usepackage{epstopdf}
\usepackage{xcolor}
\usepackage{hyperref}

\headevenname{\zihao{-5}{\textbf{\emph{Tsinghua Science and Technology, August}} 2023, 00(0): 000--000}}%
\headoddname{{\sf Sawwan et al.:}\quad {\textbf{\emph{Diversity-Based Recruitment in Crowdsensing By Combinatorial Multi-Armed Bandits}}}}%

\newtheoremstyle{mystyle}{0pt}{0pt}{\normalfont}{1em}{\bf}{}{1em}{}
\theoremstyle{mystyle}
\renewcommand\figurename{Fig.~}

\newtheorem{lemma}{\textbf{Lemma}}
\newtheorem{theorem}{\textbf{Theorem}}

\newcommand{\nop}[1]{}

\makeatletter
\renewcommand{\@biblabel}[1]{[#1]\hfill}
\makeatother
\begin{document}
\thispagestyle{empty}
\clearpage
\hyphenpenalty=50000
\makeatletter
\newcommand\mysmall{\@setfontsize\mysmall{7}{9.5}}
\newenvironment{tablehere}
  {\def\@captype{table}}
  {}
\newenvironment{figurehere}
  {\def\@captype{figure}}
  {}
\thispagestyle{plain}%
\thispagestyle{empty}%
\let\temp\footnote
\renewcommand \footnote[1]{\temp{\zihao{-5}#1}}
{}
\vspace*{-40pt}
\noindent{\zihao{5-}\textbf{\scalebox{0.885}[1.0]{\makebox[5.9cm][s]
{TSINGHUA\, SCIENCE\, AND\, TECHNOLOGY}}}}
\vskip .2mm
{\zihao{5-}
\textbf{
\hspace{-5mm}
\scalebox{1}[1.0]{\makebox[5.6cm][s]{%
I\hfill S\hfill S\hfill N\hfill{\color{white}%
l\hfill l\hfill}0\hfill0\hfill0\hfill0\hfill-\hfill0\hfill0\hfill0\hfill0
\hfill \color{white}{\quad 0\hfill ?\hfill /\hfill ?\hfill ?\quad p\hfill p\hfill  ?\hfill ?\hfill ?\hfill --\hfill ?\hfill ?\hfill ?}\hfill}}}}
\vskip .2mm
{\zihao{5-}
\textbf{
\hspace{-5mm}
\scalebox{1}[1.0]{\makebox[5.6cm][s]{%
DOI:~\hfill~\hfill0\hfill0\hfill.\hfill0\hfill0\hfill0\hfill0\hfill0\hfill/\hfill T\hfill S\hfill T\hfill.\hfill2\hfill0\hfill2\hfill3\hfill.\hfill0\hfill0\hfill0\hfill0\hfill0\hfill0\hfill0}}}}
\vskip .2mm\noindent
{\zihao{5-}\textbf{\scalebox{1}[1.0]{\makebox[5.6cm][s]{%
\color{white}{V\hfill o\hfill l\hfill u\hfill m\hfill%
e\hspace{0.356em}1,\hspace{0.356em}N\hfill u\hfill%
m\hfill b\hfill e\hfill r\hspace{0.356em}1,\hspace{0.356em}%
S\hfill e\hfill p\hfill t\hfill e\hfill%
m\hfill b\hfill e\hfil lr\hspace{0.356em}2\hfill0\hfill1\hfill8}}}}}\\

\begin{strip}
{\center
{\zihao{3}\textbf{
Diversity-Based Recruitment in Crowdsensing By Combinatorial Multi-Armed Bandits}}
\vskip 9mm}

{\center {\sf \zihao{5}
Abdalaziz Sawwan$^*$ and Jie Wu}
\vskip 5mm}

%

\centering{
\begin{tabular}{p{160mm}}

{\zihao{-5}
\linespread{1.6667} %
\noindent
\bf{Abstract:} {\sf
This paper explores mobile crowdsensing, which leverages mobile devices and their users for collective sensing tasks under the coordination of a central requester. The primary challenge here is the variability in the sensing capabilities of individual workers, which are initially unknown and must be progressively learned. In each round of task assignment, the requester selects a group of workers to handle specific tasks. This process inherently leads to task overlaps in the same round and repetitions across rounds. We propose a novel model that enhances task diversity over the rounds by dynamically adjusting the weight of tasks in each round based on their frequency of assignment. Additionally, it accommodates the variability in task completion quality caused by overlaps in the same round, which can range from the maximum individual worker's quality to the summation of qualities of all assigned workers in the overlap. A significant constraint in this process is the requester's budget, which demands an efficient strategy for worker recruitment. Our solution is to maximize the overall weighted quality of tasks completed in each round. We employ a combinatorial multi-armed bandit framework with an upper confidence bound approach for this purpose. The paper further presents a regret analysis and simulations using realistic data to demonstrate the efficacy of our model.}
}

\end{tabular}
}
\vskip 6mm

\vskip -3mm
\zihao{6}\end{strip}

\thispagestyle{plain}%
\thispagestyle{empty}%
\makeatother
\pagestyle{tstheadings}

\begin{figure}[b]
\vskip -6mm
\begin{tabular}{p{44mm}}
\toprule\\
\end{tabular}
\vskip -4.5mm
\noindent
\setlength{\tabcolsep}{1pt}
\begin{tabular}{p{1.5mm}p{79.5mm}}
$\bullet$& The authors are with The Center for Networked Computing, Temple University, Philadelphia, PA 19122, United States.
\\
&E-mail addresses: \{sawwan,jiewu\}@temple.edu.
\\
$\sf{*}$&
To whom correspondence should be addressed. \\
\end{tabular}
\end{figure}\zihao{5}

\section{Introduction}

There has been a great deal of interest in recent years in Mobile Crowdsensing (MCS) systems, which involve a requester recruiting a crowd of mobile users, equipped with their own mobile devices, to complete various sensing tasks \cite{1,2,3r,5,6r,Sawwan,8,9}. These mobile workers are able to utilize the advanced technologies available on their smartphones to perform a wide range of sensing tasks, and their mobility allows them to cover large areas for these purposes. The use of a crowd of mobile users is particularly useful for tasks that would be difficult for a single individual to accomplish, such as collecting noise information, gathering traffic data, monitoring water pollution, and so on.

One of the key benefits of MCS systems is the ability to gather data in a cost-effective manner by leveraging the resources of a large number of individuals. These systems have the potential to facilitate the collection of valuable information for a variety of applications, and the use of mobile devices allows for flexibility and convenience in the data gathering process. The requester in an MCS system has the ability to assign tasks to the mobile workers and can choose which tasks to prioritize based on their relative importance, allowing for efficient resource allocation and data collection.

In classic MCS systems, requesters of service publish the required sensing tasks and recruit mobile users, referred to as workers, through a platform hosted on the cloud. In practical settings, it is common for users to have different types of smartphones, leading to variations in the quality of sensing among workers. This motivates requesters to adopt strategies that aim to maximize sensing quality while minimizing the costs incurred by workers. The design of effective recruiting strategies is a crucial problem in MCS systems, and as a result, a significant amount of research has been dedicated to this topic \cite{4,7,10,11,12,13,14,15}. However, many of these strategies are based on the unrealistic assumption that the requester has prior knowledge of the sensing qualities of workers' devices, and do not take into account the diverse allocation of tasks. In addition, the problem of having unknown sensing qualities of workers has received relatively limited attention in the literature.

In addition, Ul Hassan \emph{et al.} \cite{16} were among the first to study this problem, seeking to maximize the ratio of completed tasks under dynamic task arrival settings. Wu \emph{et al.} \cite{17} developed an algorithm that recruits workers based on the Thompson sampling model, but with the assumption that the sensing quality of workers can vary for each task. Gao \emph{et al.} \cite{gao2022combination, gao} also considered the case of heterogeneous MCS systems with unknown workers, but their approach used Upper Confidence Bound (UCB) to model the problem, without taking into account task diversity over rounds. It is clear that there is a need for more comprehensive and realistic approaches to solving the problem of recruiting workers in MCS systems, particularly those that can handle the challenges of unknown sensing qualities and diverse task allocation.
\begin{figure}[t]
    \begin{center}
        \includegraphics[scale=0.44]{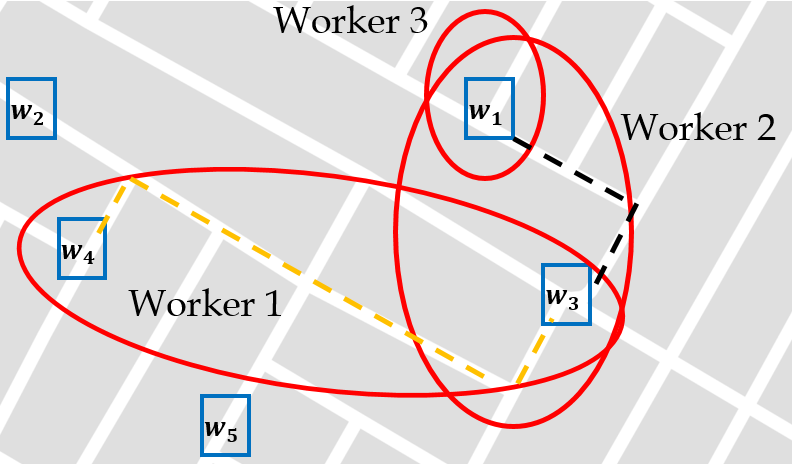}
    \end{center}
    \caption{A basic illustration of one round of the MCS system. Each one of the tasks has a weight $w_j$ that reflects its importance. Each one of the workers has a different sensing quality. Dashed lines represent trajectories for workers.}
    \label{fig1}
\end{figure}

In this paper, we consider encouraging choosing diverse tasks over the rounds as we develop an MCS recruitment strategy in which the sensing qualities of workers are being learned. Furthermore, we propose a generic expression with a tunable parameter for overlaps. This expression is seamlessly integrated into the classic multi-armed bandit model and its analysis.
Consider the case where the sensing tasks are actually taking videos or pictures of the traffic in a certain city over a period of time where the requester uses the platform to recruit some mobile users who act as the workers. Hence, we have multiple weighted tasks (weight represents importance) that need to be covered in multiple rounds. Repeated coverage of the same task is allowed, with a discounted value. Each worker has multiple possible subsets of tasks, each of which corresponds to a worker's possible daily travel trajectory. Each worker requests a cost with an uncertain quality for task coverage. The objective is to maximize overall utility, defined as a summation of coverage quality multiplied by coverage weight, constrained by the budget. Coverage weight is the weight of each task multiplied by the discounted multiple coverages of the task. Weight values diminish with repetition over the rounds to encourage diversity. Figure \ref{fig1} shows the basic illustration of the problem. Note that the number of tasks for each worker is not necessarily the same.


In this work, the requester publishes their tasks to the platform. Then, the workers submit their possible sets of tasks to cover alongside the cost values of each one of those possible sets. Afterward, over many rounds, the requester tells each of the workers which of those sets to cover or if they are not recruited, and the workers upload their pictures or videos. The sensing quality of each worker follows an unknown probability distribution to be learned over the rounds. So, our goal is to develop a strategy that maximizes the total weighted quality values constrained by a certain budget.

In our MCS framework, we address several unique challenges to optimize task allocation and worker recruitment. A key challenge is the dynamic adjustment of task weights over successive rounds. Specifically, as a task is covered in one round, its weight value is decreased for subsequent rounds. This approach is designed to prevent the recurrent selection of the same tasks, thereby promoting a diverse range of task coverage. Such a mechanism is vital to avoid data redundancy and ensure a comprehensive understanding of the sensing field. Furthermore, we address the complexity of handling overlaps in task coverage, where tasks may be sensed by multiple workers simultaneously. In this scenario, our model calculates the total completion quality based on both individual and collective worker qualities. This nuanced treatment of overlaps ensures that the data quality is not compromised despite the concurrent sensing efforts. Additionally, we consider the challenge of learning the unknown sensing qualities of workers over time. This aspect is crucial, as it directly influences the overall quality of the data collected. By integrating these challenges into our Combinatorial Multi-Armed Bandit (CMAB) framework, we propose a novel approach that significantly enhances the effectiveness of mobile crowdsensing systems.

Our new results in this work are summarized as follows:

\begin{itemize}
  \item We design a new utility model that penalizes repeated coverage in MCS in order to indirectly encourage diverse coverage over the rounds. A novel $K$-arm CMAB problem is introduced.
  \item We propose a generic expression with a tunable parameter for overlaps. This expression is seamlessly integrated into the classic multi-armed bandit model and its analysis.
  \item We run extensive simulations on real-world data in order to compare the performance of our algorithms with existing ones.
\end{itemize}

The remainder of the paper is organized as follows. In Section 2, a general overview is presented regarding the system and the novelty of it. Section 3 shows the formulation of the problem. Section 4 presents the solution in details. In Section 5, some related works are reviewed. In Section 6, simulation results are presented to evaluate the performance of our solution on real-world traces. Finally, Section 7 gives the conclusion.

\begin{figure}[t]
    \begin{center}
        \includegraphics[scale=0.38]{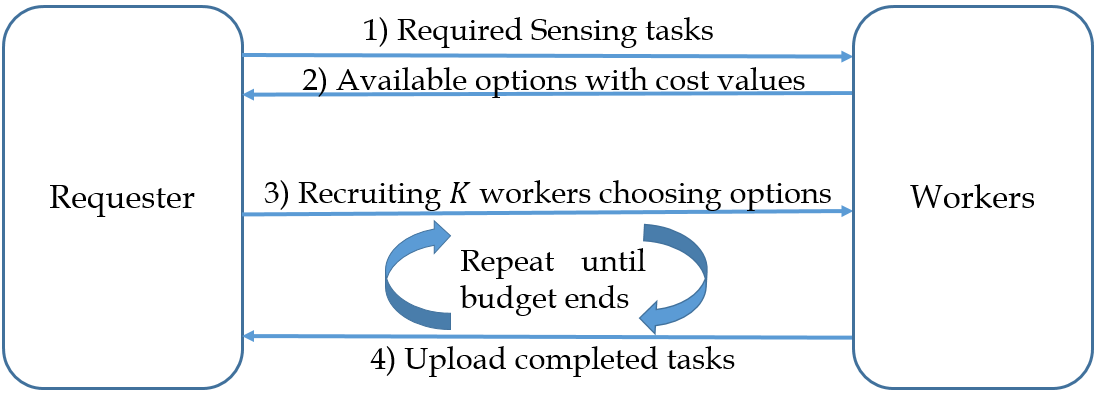}
    \end{center}
    \caption{A basic illustration of one round of the MCS system. Each one of the tasks has a weight $w_i$ that reflects its importance. Each one of the workers has a different sensing quality.}
    \label{MCS}
\end{figure}

\section{General Overview}

In this section, we formulate the problem after giving a general overview of the MCS system concerned. Now, we consider an MCS arrangement in which there is a crowd of mobile users who are ready to be employed as workers to do a sensing task, say collecting traffic information using the cameras of their mobile devices. On the other hand, there is a requester who operates on a platform in order to communicate with the workers. This requester has a limited budget to do the tasks. Each one of those tasks is fixed in its location and has a specific changing weight value that reflects its relative significance. Sensing the tasks starts with the requester that only publishes the tasks to the workers on the platform.

Afterward, those workers submit a group of subsets of tasks (called options) with the cost they will incur for each one of those options. The first round of sensing starts with the requester recruiting $K$ workers by specifying a certain previously-submitted option for each one of those workers. Then, the workers do their tasks and upload their sensing results to the platform. Then, the second round starts by choosing more $K$ workers and so on until the budget of the requester is spent. Figure \ref{MCS} shows the steps of the process.

\begin{figure}[t]
    \begin{center}
        \includegraphics[scale=0.38]{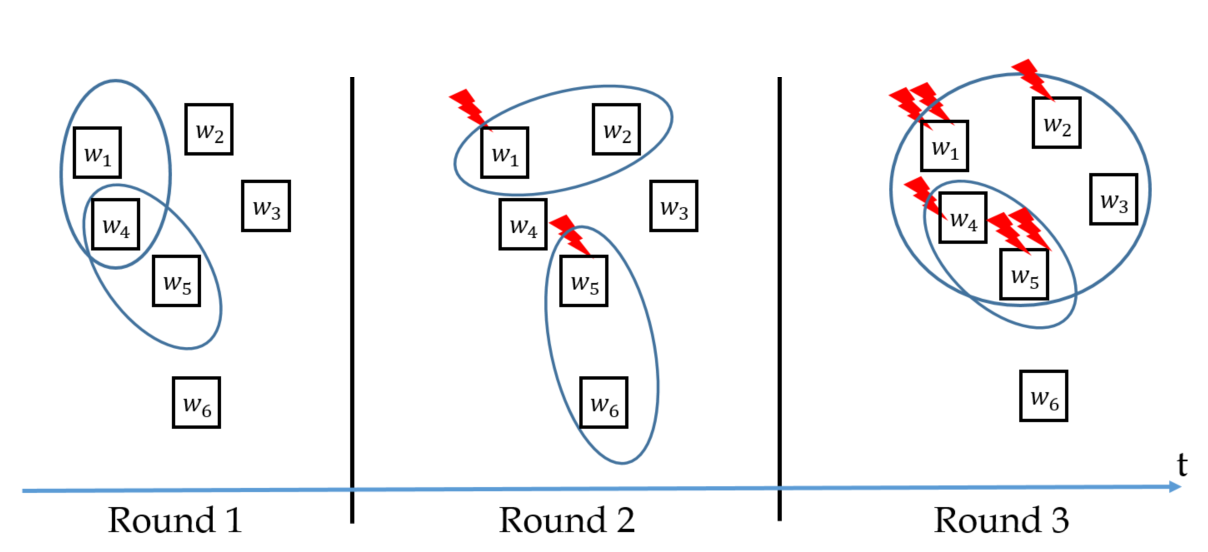}
    \end{center}
    \caption{An illustration of the novel model of diversity in the mobile crowdsensing problem. The red signs include a certain penalty incurred over the utility return from a certain task because of repetition.}
    \label{divv}
\end{figure}

\begin{figure}[t]
    \begin{center}
        \includegraphics[scale=0.40]{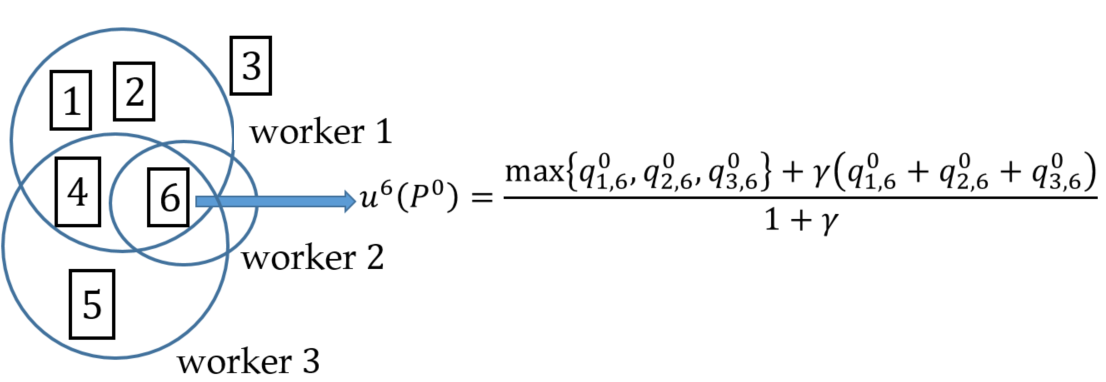}
    \end{center}
    \caption{An illustration of having more than one worker covering the same task at the same round $0$. The final completion quality of the tasks in overlaps depends on $\gamma$.}
     \label{ovv}
\end{figure}

One of the key features of our model is the emphasis on encouraging diversity in the selection of tasks over the various rounds of sensing. To achieve this, we utilize a method of penalizing repetition, which decreases the weight of a task chosen by a certain amount for each round in which it has been previously selected. This approach is designed to promote the diversity of task selection and is demonstrated in Figure \ref{divv}. In addition, tasks that are covered by multiple workers at the same time, referred to as overlaps, return a value that is dependent on the sensing qualities of all the covering workers. The sensing quality of each worker can range from the maximum sensing quality of the covering workers to the summation of their qualities, as illustrated in Figure \ref{ovv}. The distribution of sensing quality for each worker is not known and is learned over the course of the various rounds of sensing. It is worth noting that we make the assumption that the system is able to maintain the truthfulness, which was studied extensively by Wang \emph{et al.} \cite{wang2022truthful} and Zhang \emph{et al.} \cite{zhang2022enabling} for crowdsensing models similar to the one we study in this work.

Furthermore, we assume the security and privacy of the crowdsensing system, and therefore do not need to consider incentives in our model. These warranted assumptions allow us to focus on the task of selecting and allocating tasks to the workers in a manner that promotes diversity and efficient resource utilization. There are a variety of frameworks that can be used to quantify the diversity of a chosen set, such as entropy, but our model focuses on penalizing repetition as a means of encouraging diversity.

\section{Problem Formulation}

In this section, we show the preliminaries needed before formulating the problem.

\subsection{Preliminaries}

We will use a similar standard notation for mobile crowdsensing systems, \cite{gao2022combination} is an example. Let the budget of the requester be $B$, $\mathcal{N}$ represents the set of workers $\{1,\dots, N \}$, $\mathcal{M}$ represents the set of all tasks $\{1,\dots, M \}$, and $t$ is the number of the round. The weight of task $j$ at round $t$ is denoted by $w_j^t$. Without loss of generality, we can start initially with normalized weight values $\Sigma_{j\in\mathcal{M}}w_j^1=1$. Normalizing weight values helps us make sense of the relative value for each task in comparison to the total summation of weights of all tasks.

The present system employs a variable, $L$, to represent the greatest number of options that any worker on the platform has submitted. For any given worker, $i$, the $l$-th option among the available options, where $1\leq l\leq L$, is represented by $p^l_i$. This variable, $p^l_i$, comprises a subset of tasks, designated by $\mathcal{M}^l_i$, as well as the cost associated with this particular option, denoted by $c_i^l$. It is important to note that the subscript $i$ denotes the specific worker in question, while the superscript $l$ refers to the specific option being considered among the available options for that worker.

During each round of the system, the requester is permitted to select a maximum of one option from each worker. Furthermore, it is assumed that the cost of an option, $c_i^l$, is determined by the number of tasks contained within that option, such that $c_i^l=\epsilon_i f(|\mathcal{M}^l_i|)$. In this equation, $f(.)$ represents a monotonically increasing function, while $\epsilon_i$ denotes the cost factor that differentiates the cost values incurred by each individual worker.

This model makes the ratio of cost values for two options with the same size from different workers fixed. Furthermore, this cost modeling fixes the cost values of different options of the same size for one worker. It is safe to make that assumption since we are making the cost of an option with a larger number of tasks never cheaper than an option with a smaller number of tasks (although in some special cases this assumption must be relaxed). This modeling of the cost gives the ability to let workers with more advanced devices charge more cost by choosing a proper corresponding cost factor that would reflect how advanced are their smartphones. $\mathcal{O}$ denotes the set of all options of all the workers.

Now, regarding the distributions of the sensing quality of worker $i$ (denoted by $Q_i$), we assume that it is bounded, and without loss of generality, that it gives a value in $[0,1]$. That is, $q_{i,j}^t \sim Q_i$, $q_{i,j}^t \in [0,1]$ where $Q_i$ represents the bounded distribution of the quality values given from worker $i$, and that $q_{i,j}^t$ is the resulting sample of the distribution when “pulling the arm” and choosing worker $i$ with an option that includes task $j$ at round $t$.

We make the reasonable assumption that the distributions are independent. we define $q_i=\mathbb{E}[Q_i]$, which is the average sensing quality of worker $i$, and which is what our strategies try to learn. Having this kind of probability distribution for sensing qualities of workers is realistic as the workers' sensing qualities have for the same task would be affected by the momentary condition of the worker, their skill, how they use their camera to capture a video or picture of the traffic, etc. Different from classical CMAB models \cite{21,22}, the fact each time an option $p_i^l$ is chosen, the distribution $Q_i$ of sensing quality for worker $i$ is learned many times in the same round. In other words, it will be learned $|\mathcal{M}^l_i|$ times.

One of the most significant innovations in our work is the introduction of a mechanism for dynamically adjusting the weight values $w_j$ of tasks as they are completed over successive rounds. Specifically, as the number of rounds in which task $j$ has been completed, represented by $m_j(t)$, approaches infinity, the weight value approaches a specified fraction, referred to as the diversity ratio and denoted by $\kappa$, of the initial weight value ($w_j\rightarrow \kappa w_j^1$ as $m_j(t) \rightarrow \infty$.). This ensures that the weight values of tasks do not become too heavily skewed towards tasks that have been completed less frequently, and instead promotes a more balanced distribution of tasks. This novel approach to task weighting has a number of important implications. By encouraging diversity in the selection of tasks, it helps to prevent the algorithm from becoming too heavily focused on any one particular task, which can lead to suboptimal performance. In addition, by continually adjusting the weight values of tasks as they are completed, it helps to ensure that the algorithm remains responsive to changes in the task environment and can effectively adapt to new circumstances. Overall, the implementation of this task weighting mechanism represents a significant contribution to the field and has the potential to improve the performance of a wide range of algorithms.

The other key contribution of our work is the development of a model for capturing the return on sensing an overlapped task, which is defined as a task that is being sensed by multiple workers simultaneously. This model is based on the assumption that the deterioration in the quality of the task's completion follows an exponential decay function with a rate of $\lambda>0$. To model the return of sensing an overlapped task, we propose representing the total completion quality of the task as lying within a range between the maximum individual sensing quality of the covering workers and the summation of their sensing qualities. The exact location within this range is determined by a factor called the overlapping factor, denoted by $\gamma$, which is a non-negative number. This approach allows us to accurately capture the impact of overlapping on task quality and to account for the diverse ways in which overlapped tasks may be completed. Overall, this novel approach to modeling overlapped tasks represents an important contribution to the field and has the potential to improve the performance of a wide range of algorithms.



\subsection{The Problem}

For this diversity-aware MCS system, our objective is to recruit exactly $K$ workers (choosing the specific option for each worker), in each round in a way that maximized the total weighted sum of completed sensing qualities with the consideration of overlaps over all the rounds. The budget is limited in this regard. Denote the set of all selected options at round $t$ by $\mathcal{O}^t\in \mathcal{O}$ ($p_i^l\in \mathcal{O}^t$ represents the $l$-th option of worker $i$ selected at round $t$). To indicate that exactly one option for each worker is selected, $\Sigma_{l=1}^L\mathbb{I}\{ p_i^l \in \mathcal{O}^t \}\leq 1$ for all $i$. Such that $\mathbb{I}\{ true \} = 1$ and $\mathbb{I}\{ false \} = 1$. Now, the penalizing repetition can be modeled by any monotonically-decreasing function such that $w_j^\infty=\kappa w_j^1$. We opt to choose: 
\begin{equation}
    w_j^t=((1-\kappa)e^{-m_j(t)/\lambda}+\kappa)w_j^1
    \label{eq1}
\end{equation}

\noindent such that:

\begin{equation}
        m_j^t=
        \left\{ \begin{array}{ll}
            m_j^{t-1}+1; &p_i^l\in \mathcal{O}^t \\
            m_j^{t-1}; &\text{otherwise}
        \end{array} \right.
    \label{eq1a}
    \end{equation}

On the other hand, the overlap-aware total quality for a certain task can be modeled as:

\begin{equation}
        u^j(\mathcal{O}^t)=
        \left\{ \begin{array}{ll}
            \frac{max\{ q_{i,j}^t | p_i^l \in \mathcal{O}^t \}+\gamma (\Sigma_{i|p_i^l \in \mathcal{O}^t}q_{i,j}^t)}{1+\gamma};& j\in \mathcal{O}^t  \\
            0; &\!\!\!\!\!\text{otherwise}
        \end{array} \right.
        \label{eq3}
    \end{equation}

\begin{figure}[t]
    \begin{center}
        \includegraphics[scale=0.3]{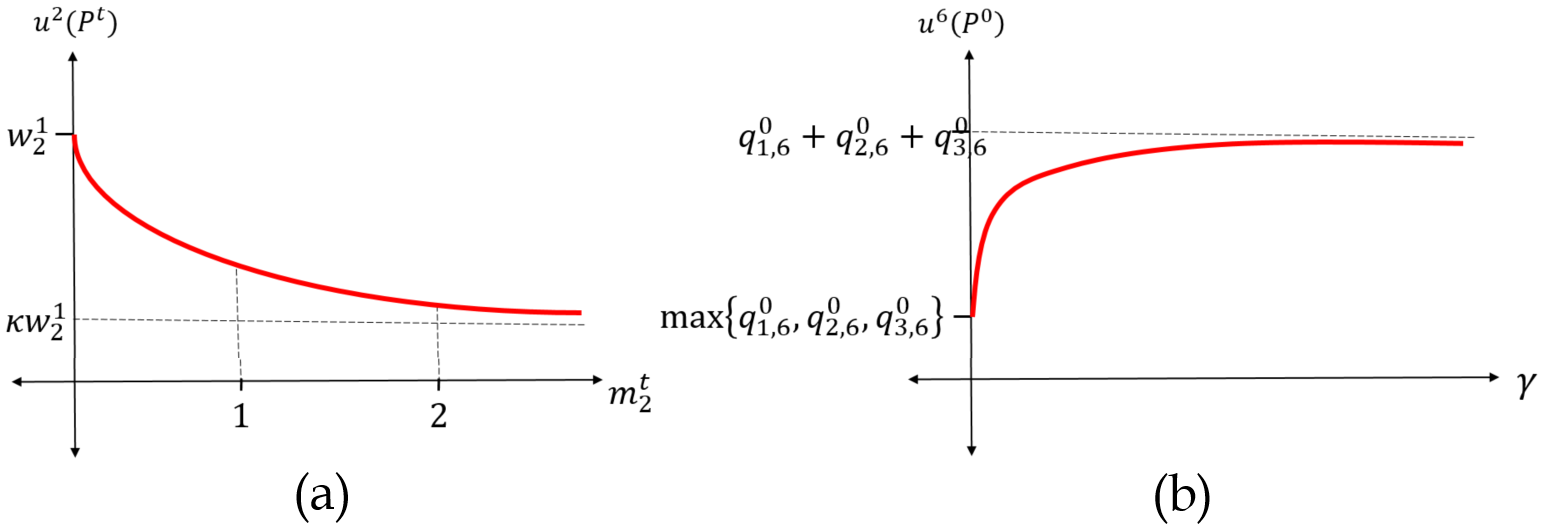}
    \end{center}
    \caption{Plots of the change of utility values in the new model. (a) shows penalizing repetition. (b) shows the overlapping factor.}
    \label{dia}
\end{figure}

Both Equations \ref{eq1} and \ref{eq3} are plotted in Figure \ref{dia}. To this end, we can define $u(\mathcal{O}^t)$ to denote the total achieved weighted completion quality of all tasks based on $\mathcal{O}^t$ in round $t$ to be:

\begin{equation}
    u(\mathcal{O}^t)=\Sigma_{j\in \mathcal{M}}(w_j^t\times u^j(\mathcal{O}^t)).
    \label{wcq}
\end{equation}

To see it more clearly, our goal in this work is to determine the set $\{ \mathcal{O}^t|\forall t \in [1,\tau(B)] \}$, where $\tau(B)$ is the round at which the budget becomes so small so no more rounds can be done. This set needs to ensure that the expected total  weighted completion quality shown in Equation \ref{wcq} is maximized under the constraints. More formally:
\begin{align}
    \text{maximize} \: &\mathbb{E}[\Sigma_{t}u(\mathcal{O}^t)] \label{eqa} \\
    \text{s. t.} \; \; \; \; &\Sigma_{\forall p_i^l \in \mathcal{O}^t} c_i^l \leq B \label{eqb} \\
    &\Sigma_{\forall l}\mathbb{I}\{ p_i^l \in \mathcal{O}^t \} \leq 1 \label{eqc} \\
    &|\mathcal{O}^t|=K \; \; \; \; \forall t \label{eqd}
\end{align}
In other words, Equation \ref{eqa} refers to the main objective of maximizing the total expected weighted quality values over the rounds. Equation \ref{eqb} refers to the budget constraint. Equation \ref{eqc} refers that at most one option of each worker is chosen every round. finally, the constraint in Equation \ref{eqd} represents that exactly $K$ workers must be deployed every round.

\section{Solution of the Problem}\label{sec:Solution}

In this section, we introduce the algorithm that solves the novel problem considering both diversity and overlaps.

\subsection{Algorithm Overview}
The problem at hand is significantly more complex and challenging to analyze compared to previous work in this area, particularly the basic work of Gao \emph{et al.} \cite{gao}. This added complexity is due to the inclusion of two additional layers of considerations beyond the fundamental elements of the problem. The first layer pertains to the handling of overlaps or repeated coverage, which is an important factor to consider when determining the optimal allocation of tasks among the workers. This aspect of the problem is described in detail in Equation \ref{eq3}. The second layer involves the modeling of diversity through the penalization of repetition of tasks over time, as outlined in Equation \ref{eq1}. This aspect of the problem is critical for ensuring that the selection of tasks is diverse and well-balanced, rather than overly concentrated on a small number of tasks.

In order to address the issue of having unknown quality distributions for the workers $Q_i$, we utilize an extended $K$-CMAB model that is constrained by a budget. This model allows us to consider the trade-off between the cost of recruiting workers and the benefits of their sensing capabilities. Each worker is treated as an arm in this model, and the selection of a worker with an option corresponds to the selection of an arm and a specific option within that arm. The completion quality of an option for a worker is analogous to the reward returned by an arm. By choosing workers and options that provide the highest expected reward given the budget constraints, we can maximize the overall utility of the system. During each round, exactly $K$ workers are recruited, providing the requester with multiple opportunities to learn the sensing quality of each of these workers and adjust their selection strategies accordingly.

In order to do the learning of those quality values, we employ the classical solution of the exploration versus exploitation dichotomy, which is the UCB method \cite{21}. We extend the method to accommodate our application. Afterward, we introduce the estimation function of worker qualities that is based on UCB with the taking of the optimization problem in Equations \ref{eqa}-\ref{eqd}. Then, a simple greedy algorithm that chooses $K$ workers each round with the maximum ratio of the estimation quality function to the recruitment cost of the option. Lastly, we show that this greedy algorithm produces a bounded solution in comparison to the optimal one.

\subsection{Algorithm Design}

Since we assume that, in any round, the requester can force the worker to perform the sensing tasks submitted earlier into the platform as an available option, the worker's quality gets learned by the number of tasks covered by that chosen option $|\mathcal{M}_i^l|$. This raises the need to count how many times the quality of worker $i$ was learned, and how many times a specific option $l$ was chosen. 

\begin{equation}
        n_i^l(t)=
        \left\{ \begin{array}{ll}
            n_i^l(t)+1; &p_i^l\in \mathcal{O}^t \\
            n_i^l(t-1); &\text{otherwise}
        \end{array} \right.
    \label{eqcount1}
    \end{equation}
    
\begin{equation}
    n_i(t)=\Sigma_{l=1}^{L} ((n_i^l(t)-n_i^l(t-1)) \times |\mathcal{M}_i^l|)
    \label{eqcount2}
\end{equation}

Equations \ref{eqcount1}-\ref{eqcount2} show the record of the times a worker was recruited, and the number of times an option was chosen, at a specific round $t$.

We now define the average of the measured quality values of worker $i$ at round $t$. We denote this measured average by $\bar{q}_i(t)$. This average simply equals the following:

\begin{equation}
        \bar{q}_i(t)=
        \left\{ \begin{array}{ll}
            \frac{\bar{q}_i(t-1)n_i(t-1)+\Sigma_{j\in \mathcal{M}_i^l} q_{i,j}^t}{n_i(t-1)+|\mathcal{M}_i^l|}; &j\in \mathcal{O}^t  \\
            \bar{q}_i(t-1); &\text{otherwise}
        \end{array} \right.
        \label{eqavvvv}
    \end{equation}

Now, we introduce the classic related UCB equation that balances exploration and exploitation, which is diversity-driven. We use $\hat{q}_i(t)$ to represent this UCB value for quality for worker $i$. Equation \ref{eqdef} shows it:

\begin{equation}
\hat{q}_i(t)=\bar{q}_i(t)+\sqrt{\frac{(K+1)\log{ \Sigma_{\forall i} n_i(t)}}{n_i(t)}}
    \label{eqdef}
\end{equation}

To introduce the diversity-driven and overlap-aware total completion weighted quality function that is based on the UCB, we define $U(\mathcal{O}^t)$ to denote it at round $t$ in case the set of options $\mathcal{O}^l$ is chosen. That total UCB quality function is shown in Equations \ref{eqtot} -\ref{eqtot2}.

\begin{equation}
U(\mathcal{O}^t)=\Sigma_{\forall j} w_j \hat{u}^j(\mathcal{O}^t)
    \label{eqtot}
\end{equation}

\begin{center}
\includegraphics[scale=0.55]{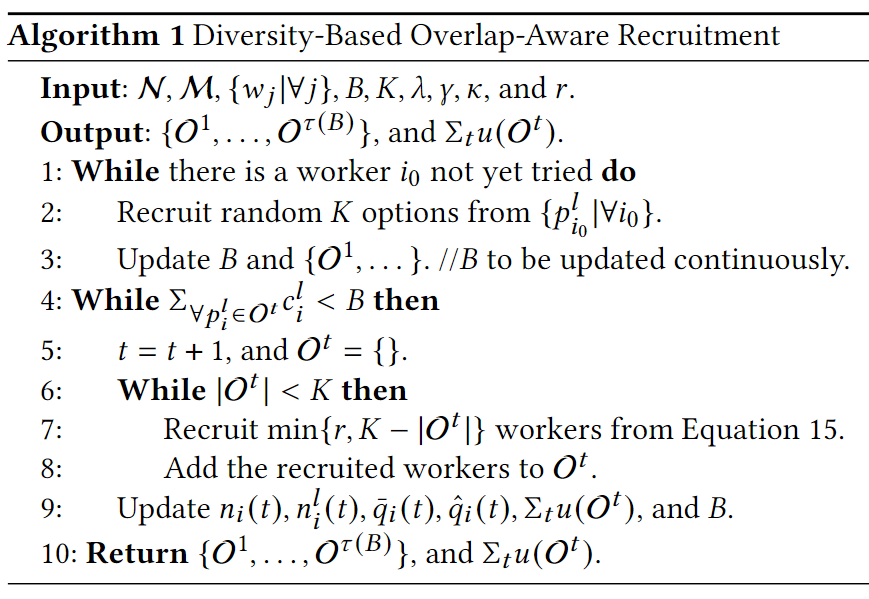}
\end{center}

\begin{equation} \hat{u}^j(\mathcal{O}^t)=\left\{ \begin{array}{ll}
            \frac{max\{ \hat{q}_{i,j}^t | p_i^l \in \mathcal{O}^t \}+\gamma (\Sigma_{i|p_i^l \in \mathcal{O}^t}\hat{q}_{i,j}^t)}{1+\gamma}; &j\in \mathcal{O}^t\\
            0; &\!\!\!\!\!\text{otherwise}
        \end{array} \right.    \label{eqtot2}
\end{equation}

This means that our greedy algorithm will simply pick the $K$ workers that maximize the total $U(\mathcal{O}^t)$ value. In more detail, we start by setting a certain accuracy parameter $r$ ($1\leq r \leq K$) that reflects how accurate $U(\mathcal{O}^t)$ will be evaluated. This parameter determines the maximum number of workers that the algorithm can consider to contribute in an overlap. When we set this parameter to $1$, then the algorithm is not overlap-aware anymore. The default value would be $2$, and choosing $K$ means that all possible overlaps are considered, but this would require enormous computational power.

When running the algorithm the algorithm will initialize the values of the counters and UCB quality estimations by selecting any option for each worker at the beginning rounds. Then, for each round afterward, the algorithm starts filling up $\mathcal{O}^t$. This is done simply as follows:

\begin{enumerate}
    \item Start with empty set $\mathcal{O}^t$.
    \item Find the next $\min \{r, K-|\mathcal{O}^t| \}$ options that would maximize $U_r(\mathcal{O}^t)$ more quickly with total cost.
    \item Repeat the previous step until choosing $K$ workers.
\end{enumerate}

\noindent Denoting $U_r(\mathcal{O}^t)$ to represent the value of $U(\mathcal{O}^t)$ evaluated with an accuracy determined by $r$ (considering overlaps with maximum of $r$ contributors).

In other words, our criterion for selection is simply the ratio between the marginal increase in the value of $U_r(\mathcal{O}^t)$ and the cost needed for that increase. More formally, the next options chosen to fill $\mathcal{O}^t$ are:

\begin{equation}
\begin{gathered}
\mathcal{O}_{\!\!next}^t\!\!=\!\!\{ p_i^l | \:p_i^l \!\!=\! \arg \!\!\!\!\!\!\!\!\max_{\mathcal{P'}_{next}^t \in \mathcal{O}/\mathcal{O}^t} {\!\!\!\!\!\!\frac{U_r(\mathcal{O}^t \cup \mathcal{P'}_{next}^t) - U_r(\mathcal{O}^t)} {\Sigma_{(i,j)\in \mathcal{P'}_{next}^t}c_i^j}},\\ 
\: \: \: \: \: \: \: \: \text{$|\mathcal{P'}_{next}^t| = \min \{r, K-|\mathcal{O}^t|\} \}$} 
\; \; \; \; \; \; \; \; \;
\end{gathered} \label{cri} \end{equation}

Notice that in each round, the number of options from different workers chosen is $\min \{r, K-|\mathcal{O}^t|\}$. After depending on Equation \ref{cri} to select the $K$ workers at round $t$, each worker does their specified set of tasks represented by $\mathcal{M}_i^l$, then, all the individual completion qualities $q_{i,j}^t$ are reported to the platform so that the record for each worker is updated. That includes all the recorded averages and UCB values $\bar{q}_i(t)$ and $\hat{q}_i(t)$, alongside the counter values $n_i(t)$ and $n_i^l(t)$, that are used repeatedly all throughout the process. On the other hand, the total accumulated qualities $\Sigma_{t}u(\mathcal{O}^t)$ is updated. The algorithm keeps iterating until the budget runs out. Algorithm 1 shows the detailed algorithm.
\subsection{Algorithm Details and Complexity}

The initial lines of the algorithm are designed to initialize various counters and sets that will be used throughout the process. The algorithm then enters a series of rounds, which are executed according to the instructions in lines 3-9. During each round, the algorithm aims to identify the optimal combination of $min {r, K-r }$ workers based on the criterion outlined in Equation \ref{cri}. This criterion is meant to ensure that the algorithm is aware of the overlap of up to $r$ workers and seeks to maximize the ratio of marginal gain in total utility to the cost of that gain. The algorithm will continue to run until the budget constraint is reached, at which point it will terminate.

It is important to note that each round $t$ requires workers to report their individual sensing quality values $q_{i,j}^t$, which are used by the platform to update the corresponding counters and measures for each worker. By utilizing this information, the platform is able to keep track of the performance and progress of each worker, allowing it to make more informed decisions about task allocation and resource utilization. Overall, the algorithm is designed to efficiently select the best combination of $min {r, K-r }$ workers at each round.

Regarding the time complexity of the algorithm, most of the computation is spent in line 7 trying to figure out the solution of Equation \ref{cri}. Disregarding the initialization, that would take a time of $O(\lceil N/K\rceil)$ to finish. This denomination of line 7 with the loop yield directly to the total time complexity of $O(KL^rN^rM^r)$.

\subsection{Bound Performance}

If we make the assumption that the distributions of the sensing qualities for each worker $Q_i$ are known and are independent and identically distributed (i.i.d.), the problem can be reduced to a special 0-1 knapsack problem if we set $\gamma=0$ and $\kappa=1$. This special case is a well-known NP-hard problem \cite{23}, meaning that there is no known polynomial-time algorithm that can solve it optimally. However, our proposed algorithm provides an approximate solution that carefully selects workers according to the criterion outlined in Equation \ref{cri}. This approximate solution is denoted by $\mathcal{O}^\star$. Despite the fact that it is not a perfect solution, $\mathcal{O}^\star$ still provides satisfactory performance in practice, as it is able to effectively balance the trade-off between maximizing the total utility and minimizing the cost.

This solution always holds that $\hat{u}^j(\mathcal{O}^t)\geq \alpha \times \max_{\mathcal{O}^t \in \mathcal{O}} \hat{u}^j(\mathcal{O}^t)$, such that $\alpha$ is the approximation factor ($0\leq \alpha \leq  1$). The definition of the overlap-aware $\hat{u}^j(\mathcal{O}^t)$ can be found in Equation \ref{eqtot2}.

In addition, since we are trying to bound over the $\alpha-$approximation of the optimal solution, we do the $\alpha-$approximation regret analysis \cite{21, 24} instead of the traditional one. This means that the regret under a certain budget $B$ becomes defined as in Equations \ref{eqreg}-\ref{eqreg2}:
\begin{equation}
R_\alpha(B)=\alpha \times \Sigma_{t}u(\mathcal{O}^*) - \mathbb{E}[\Sigma_{t}u(\mathcal{O}^t)] \label{eqreg}
\end{equation}
\begin{equation}
R_\alpha(B)\leq \Sigma_{t}u(\mathcal{O}^\star) - \mathbb{E}[\Sigma_{t}u(\mathcal{O}^t)] \label{eqreg2}
\end{equation}

Following the common convention, we denote the optimal workers with superscript $*$, while the $\alpha$-optimal workers with the $\star$ sign. Now, considering the diversity-based overlap-aware criterion shown in Equation \ref{cri}, we know beforehand that $\alpha$ for our algorithm can not be less than $0.5$ \cite{23, 25}. We now can define both the largest and smallest difference possible in the values of the sensing quality of the workers who are not $\alpha$-optimal. Equations \ref{dd1}-\ref{dd2} show that:

\begin{equation}
    \Delta_{m i n}=u(\mathcal{O}^\star)-\max _{\mathcal{O}^{\prime} \neq \mathcal{O}^{\star}} u\left(\mathcal{O}^{\prime}\right)
    \label{dd1}
\end{equation}

\begin{equation}
    \Delta_{m a x}=u(\mathcal{O}^\star)-\min _{\mathcal{O}^{\prime} \neq \mathcal{O}^{\star}} u\left(\mathcal{O}^{\prime}\right)
    \label{dd2}
\end{equation}

Afterward, we need to define a new counter $C_i^l(t)$ for each one of the options. Those counters are just for the options that are chosen when a set of $K$ non-$\alpha$-optimal workers is chosen. This counter increases by one every time a non-$\alpha$-optimal set is chosen and the option $l$ from worker $i$ has the least counter value among the other options invoked. That will update this least option as $C_{i}^{l}(t)=C_{i}^{l}(t-1)+1$. The expected count of this counter is specifically bounded as in Lemma \ref{lem1}.

\begin{lemma} \label{lem1}
In the system, it holds always that $\mathbb{E}[C_i^l(\tau(B))]\leq \frac{4K^2(K+1)}{(\Delta_{min}\times c_{min})^2}\ln(NM\tau(B))+1+K\pi^2/3$.
\end{lemma}

\begin{IEEEproof}
The proof of this lemma can, in fact, be derived from previous classic work on CMAB as done in \cite{gao}. Starting by defining an indicator that specifies whether $C_{i}^{l}(t)$ changes at round $t$ or not.


\begin{equation}
\begin{aligned}
C_{i}^{l}(\tau)&=\sum_{\forall t} \mathbb{I}\left\{I_{i}^{l}(t)=true\right\} \\
&=x+\sum_{\forall t} \mathbb{I}\left\{I_{i}^{l}(t)=true, C_{i}^{l}(t) \geq x\right\} \\
&\!\!\!\!\!\!\!\!\!\!\!\!\!\!\!\leq x+\sum_{\forall t} \mathbb{I}  \left\{  u^j(\mathcal{O}^{t+1})\geq u^j(\mathcal{O}^{\star}), C_i^l(t)\geq x \right\}=x+\\
\sum_{\forall t}\!\! \mathbb{I}&\left\{\!\!\sum_{p_{i}^{l} \in \mathcal{O}^{t+1}} \!\!\!\!\mu_{i}^{l}(t\!+\!1) \frac{\hat{q}_{i}(t)}{c_{i}^{l}}\!\! \geq\!\!\!\! \sum_{p_{i}^{l} \in \mathcal{O}^{*}}\!\!\!\! \mu_{i}^{l}(\star) \frac{\hat{q}_{i}(t)}{c_{i}^{l}}, C_{i}^{l}(t) \!\!\geq \!\!x\!\!\right\}\\
\end{aligned}
\label{eq22}
\end{equation}

\noindent such that $\mu_{i}^{l}(t+1)$ represents the total product of both the effective number of sensing tasks worker $p_{i}^{l} \in \mathcal{O}^{t+1}$ is able to give and the total weight of the corresponding sensing tasks. In other words, it can be shown in Equation \ref{eq22.5}.
\begin{equation}
\mu^l_i(t+1)=\Sigma_{\forall j} \mathbb{I} \{  p_i^l=\arg \max_{p^{l'}_{i'} \in \mathcal{O}^{t+1}} {\hat{q}_{i',j}(t+1)}  \} \times w_j, \label{eq22.5}
\end{equation}
\noindent and we can easily observe that $\mu_{i}^{l}(t+1) \leq \sum_{j \in \mathcal{M}_{i}^{l}} w_{j} \leq 1$.

Now, we can use the Chernoff-Hoeffding bound to deduce the fact that $x$ is always lower-bounded as shown in the inequality in Equation \ref{eq28.5}:

\begin{equation}
\frac{4(K+1) K^{2}}{\left(\Delta_{\min } c_{\min }\right)^{2}} \ln (N M \tau(B)) \leq x \label{eq28.5}
\end{equation}

To this end, we substitute Equation \ref{eq28.5} in Equation \ref{eq22} with applying the results of the Chernoff-Hoeffding bound from \cite{20,gao}, we end up with:


\begin{equation}
\begin{aligned}
C_{i}^{l}(\tau) &\!\! \leq\!\!\left\lceil\frac{4(K+1) K^{2}}{\left(\Delta_{\min } c_{\min }\right)^{2}} \ln (N M \tau(B))\right\rceil\!\!+\!\!\Sigma_{\forall t} 2 K t^{-2} \\
& \leq \frac{4(K+1) K^{2}}{\left(\Delta_{\min } c_{\min }\right)^{2}} \ln (N M \tau(B))+1+\frac{K \pi^{2}}{3},
\end{aligned} \label{eq29} \end{equation}

\noindent which concludes the proof. \end{IEEEproof}
\vspace{+4mm}
\begin{lemma} \label{lem2}
The algorithm terminates at time $\tau(B)$ that is bounded as follows:
\begin{equation}
\tau(B) \leq \frac{2B}{c^{\star}}+\zeta_4, \end{equation}
\begin{equation}
\tau(B) \geq \frac{B}{c^{\star}}-\zeta_3-1-\frac{\zeta_1\times \zeta_3}{\zeta_2} \ln(\frac{2B}{c^{\star}}+\zeta_4).\end{equation}

where $\left\{\begin{array}{l} c_{min}=\min\{c_i^l | \forall (i,l) \}, \: c^{\star}=\sum_{p_{i}^{l} \in \mathcal{O}^{\star}} c_{i}^{l}, \\ c_{max}\!\!=\!\!\max\{c_i^l | \forall (i,l) \}, \: \zeta_{1}\!\!=\!\!\frac{4(K+1) K^{2}}{\left(\Delta_{\min } c_{\min }\right)^{2}},\\ \zeta_{2}=\ln (N M) \zeta_{1}+1+\frac{K \pi^{2}}{3}, \\ \zeta_{3}=\frac{N L \zeta_{2}}{c^{\star}}, \\ \zeta_{4}=\frac{2 N L}{K c_{\min }}\left(\zeta_{1} \ln \left(\frac{2 N L \zeta_{1}}{K c_{\min }}\right)-\zeta_{1}+\zeta_{2}\right).\end{array}\right.$
\label{lem2} \end{lemma}

\begin{IEEEproof}
First, we present the bound for the $\alpha$-optimal time $\tau^\star(B)$. Since the set of $K$ workers who are $\alpha$-optimal is already determined, we can conclude that $\tau^\star(B)=\lfloor B/c^\star\rfloor$. In other words, $\frac{B}{c^\star}-1\leq \tau^\star(B) \leq \frac{B}{c^\star}$.

Then, based on the fact that $\forall x > 0$, $x>\ln{x}+1$, we get:

\begin{equation}
\ln \tau(B) \leq \frac{K c_{\min }}{2 N L \zeta_{1}} \tau(B)+\ln \left(\frac{2 N L \zeta_{1}}{K c_{\min }}\right)-1 \label{eq29.5} \end{equation}

Now, we start from the upper-bound of $\tau(B)$:

\begin{equation}
\begin{gathered}
\tau(B) \leq \tau^{\star}(B)+\tau(\Sigma_{p_{i}^{l} \notin \mathcal{O}^{\star}} n_{i}^{l}(\tau(B)) c_{\max }) \\
\leq \tau^{\star}(B)+N L /(K c_{\min }) \mathbb{E}[C_{i}^{l}(\tau(B))],
\end{gathered} \label{eq30} \end{equation}

Now, to prove the lower bound, we use the same previous notation to use $B^\star$ as the total budget spent on the $\alpha$-optimal workers. We use $B^-$ to represent the total budget spent on the workers who are not $\alpha$-optimal. The two values add up to $B$. From there, we have:

\begin{equation}
\tau(B)=\tau\left(B^{\star}+B^{-}\right) \geq \tau\left(B^{\star}\right) \geq \tau^{\star}\left(B^{\star}\right) \end{equation}

\begin{equation}
\begin{aligned}
&\geq \tau^{\star}(B-\Sigma_{p_{i}^{l} \notin \mathcal{O} \star} n_{i}^{l}(\tau(B)) c_{\max }) \\
&\geq \frac{B-N L \mathbb{E}\left[C_{i}^{l}(\tau(B))\right]}{c^{\star}}-1. \label{eq31} 
\end{aligned}
\end{equation}
Afterwards, we get:
\vspace{-3mm}

$$\begin{aligned}
&\tau(B) \leq\tau^{\star}(B)\\
&+\frac{N L}{K c_{\min }}(\zeta_{1}(\frac{K c_{\min }}{2 N L \zeta_{1}} \tau(B)+\ln (\frac{2 N L \zeta_{1}}{K c_{\min }})-1)+\zeta_{2}) \\
&\leq \frac{B}{c^{\star}}+\frac{\tau(B)}{2}+\frac{N L}{K c_{\min }}(\zeta_{1} \ln (\frac{2 N L \zeta_{1}}{K c_{\min }})-\zeta_{1}+\zeta_{2}) \\
&\leq \frac{2 B}{c^{\star}}+\frac{2 N L}{K c_{\min }}(\zeta_{1} \ln (\frac{2 N L \zeta_{1}}{K c_{\min }})-\zeta_{1}+\zeta_{2})=\!\!\frac{2 B}{c^{\star}}\!\!+\!\!\zeta_{4}
\end{aligned}$$
Finally, we substitute Equation \ref{eq31} to get:
$$\begin{aligned}
&\tau(B) \geq B / c^{\star}-N L \zeta_{2} / c^{\star}-1-N L \zeta_{1} \ln (\tau(B)) / c^{\star} \\
&\geq B / c^{\star}-N L \zeta_{2} / c^{\star}-1-N L \zeta_{1} \ln (2 B / c^{\star}+\zeta_{4}) / c^{\star} \\
&=B / c^{\star}-\zeta_{3}-1-\ln (2 B / c^{\star}+\zeta_{4}) \zeta_{1} \zeta_{3} / \zeta_{2} \; \; \; \; \; \; \; \; \; \; \; \; \; \;  \; \; \; \; \; \; \; \; \; \; \; \qedhere \end{aligned}$$

\end{IEEEproof}
To this end, we will have the following theorem.
\setcounter{theorem}{0}
\begin{theorem}

The upper bound of the $\alpha-$approximation regret of our greedy algorithm is $O\left(N L K^{3} \ln B\right)$, in details,

$R_{\alpha}(B) \leq\left(N L \Delta_{\max } \zeta_{1}+u^{\star} \zeta_{1} \zeta_{3} / \zeta_{2}\right)\left(\ln \left(\frac{2 B}{c^{\star}}+\zeta_{4}\right)\right)+\zeta_{5}$, where $\zeta_{5}=N L \Delta_{\max }+u^{\star}\left(1 / c^{\star}+\zeta_{3}+1\right)/\kappa.$ \end{theorem}
\begin{IEEEproof}
Although the asymptotic bound for the regret of our algorithm is the same as the recruitment algorithm without considering the diversity, the exact bound of ours is much different and reflects the diversity factor $\kappa$. First, we start with the fact that both Lemma \ref{lem1} and Lemma \ref{lem2} yield:

$$\begin{aligned}
&R_{\alpha}(B) \leq \Sigma_{\forall t} u(\mathcal{O}^{\star})\!\!-\!\!\mathbb{E}[\Sigma_{\forall t} u(\mathcal{O}^{t})] \\ 
&\leq \frac{(B+1) u^{\star}}{c^{\star} \kappa}-\frac{\tau(B) u^{\star}}{\kappa}+\tau(B) u^{\star}-\mathbb{E}[\Sigma_{\forall t} u(\mathcal{O}^{t})] \\
&\leq \frac{u^{\star}}{\kappa}(\frac{B+1}{c^{\star}}-\tau(B))+\sum_{i \in \mathcal{N}} \sum_{l=1}^{L} C_{i}^{l}(\tau(B)) \Delta_{\max } \\
&\leq \frac{u^{\star}}{\kappa}(\frac{B+1}{c^{\star}}-(\frac{B}{c^{\star}}-\zeta_{3}-1-\frac{\zeta_{1} \zeta_{3}}{\zeta_{2}} \ln (\frac{2 B}{c^{\star}}+\zeta_{4})) \\
&+N L \Delta_{\max }(\zeta_{1} \ln (\frac{2 B}{c^{\star}}+\zeta_{4})+\zeta_{2}) \\
&=(N L \Delta_{\max } \zeta_{1}+u^{\star} \zeta_{1} \zeta_{3} / \kappa\zeta_{2})(\ln (\frac{2 B}{c^{\star}}+\zeta_{4}))+\zeta_{5} \\
&=O(N L K^{3} \ln B). \; \; \; \; \; \; \; \; \; \; \; \; \; \; \; \; \; \; \; \; \; \; \; \; \; \; \; \; \; \; \; \; \; \; \; \; \; \; \; \; \; \; \; \; \; \; \; \; \; \; \; \; \; \; \; \; \; \qedhere \end{aligned}$$

\end{IEEEproof}

\begin{figure*}[t]
    \begin{center}
        \includegraphics[scale=0.59]{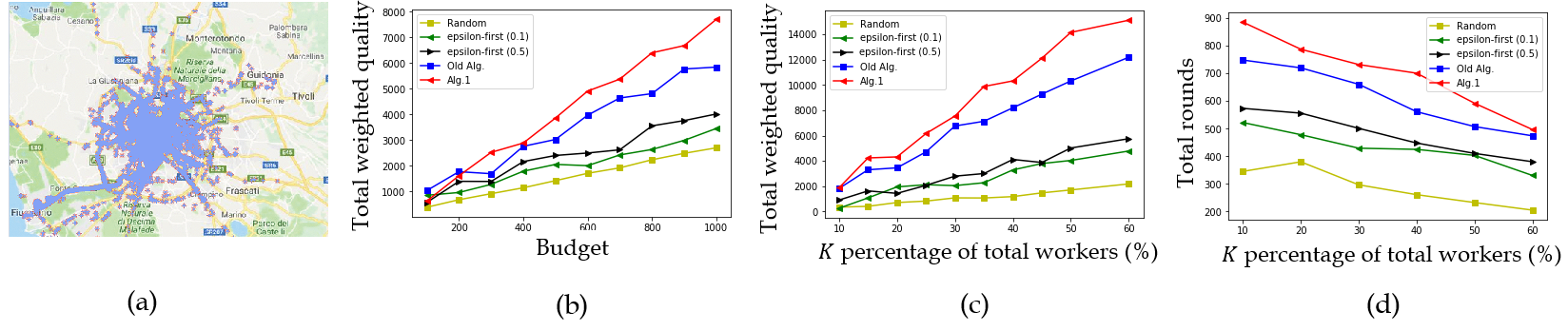}
    \end{center}
    \caption{Simulation results of the basic metrics for default values for parameters. (a) represents the locations in the traces for one of the days. (b) shows the total weighted quality as a function of the total budget. (c) shows the total weighted quality as a function of the percentage of total workers. (d) plots the number of required rounds in terms of $K$.}
    \label{s1}
\end{figure*}

\begin{figure}[t]
    \begin{center}
        \includegraphics[scale=0.63]{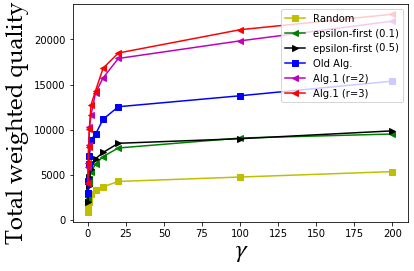}
    \end{center}
    \caption{Simulation results for the total weighted quality with varying $\gamma$ for overlap-awareness.}
    \label{s3}
\end{figure}

\section{Related Work}\label{sec:relatedwork}
The development of worker recruitment strategies in MCS systems has attracted many researchers in recent years \cite{28, trtr, 300}. However, most of this work has the typical assumption that the cost values and sensing qualities are not unknown so they try to minimize the cost or maximize the sensing qualities with some constraints. On the other hand, relatively-few work has been done on the case where the sensing values are not known and need to be learned under certain constraints \cite{16,17,24,29r,31,trtr,33,34} and none of them consider the factor of diversity or the general modeling of tasks that are covered by more than one worker in the same round.

Han \emph{et al.} \cite{33} developed an algorithm that maximizes the total reward of sensing tasks in MCS models with the constraint of a limited budget. Yang \emph{et al.} \cite{24} studied the case where the costs of workers are unknown in budgeted MCS systems too. However, both of them either make the assumption that the available options of workers are the same or that the tasks don't differ in terms of significance and don't consider any incentive to make their strategy diversify the set of chosen tasks over the rounds. On the other hand, Karaliopoulos \emph{et al.} \cite{trtr} studied the problem as a two-dimensional matching problem between the tasks and the workers but without the consideration of different weight values for the tasks.

Yang \emph{et al.} \cite{24} and Song and Jin \emph{et al.} \cite{neww3} present models in mobile crowdsensing that focus on maximizing the overall sensing quality of tasks. Yang \emph{et al.} approach this through selecting the most informative contributors within a budget, while Song and Jin aim to minimize entropy in task selection using a CMAB approach. Both models, while effective in ensuring a diverse range of tasks within a single round, do not explicitly address the challenge of promoting diversity over multiple rounds. Their approaches implicitly encourage a spread of task selection in each round but may lead to repetitive task selections across successive rounds. In contrast, our model introduces a systematic mechanism to actively encourage round-to-round task diversity. By systematically reducing the weight of tasks each time they are selected, our approach ensures a continually evolving pattern of task selection, addressing a gap not explicitly considered in the models by Yang \emph{et al.} and Song and Jin.

Zhou \emph{et al.} \cite{37} developed a UCB-based for the $K$-CMAB problem but without the consideration of having multiple options for each arm and without the consideration of having overlaps between the tasks that those options model. On the other hand, although Gao \emph{et al.} \cite{gao} have developed a novel work that is similar to ours to a certain degree where their CMAB model covers the unknown worker recruitment problem with unknown quality values and flexibility in choosing tasks with different weight values, they do not consider the case where diversity of the chosen tasks over the rounds is encouraged. Furthermore, they consider a single way to model the tasks that are covered by more than one worker at the same time. Our work, in contrast, solves the problem with changing task weight values that reflect a diversity measure while addressing a more general model for overlaps.

\section{Simulation}

In this section, we demonstrate the performance of our solution in comparison with other methods that may be used for the problem.

\subsection{Experimental Settings}

The CRAWDAD dataset \cite{26} is a valuable resource for researchers and practitioners in a variety of fields, as it provides real-world trace data that can be utilized for simulations and analysis. In our particular simulation, we utilize this dataset to evaluate the performance of our algorithm and model. The dataset consists of the locations of approximately $320$ taxi cars, which were tracked over a period of $30$ days in Rome, Italy. Figure \ref{s1} (a) provides a visual representation of the locations of the traces on one day of data collection \cite{gao}, and we have chosen to focus on a subset of $600$ of these locations for the purposes of our simulation. It is important to consider the relevance and reliability of the data being used in any simulation or analysis, and the CRAWDAD dataset has proven to be a reliable and widely-used source of trace data.

\begin{figure*}[t]
    \begin{center}
        \includegraphics[scale=0.59]{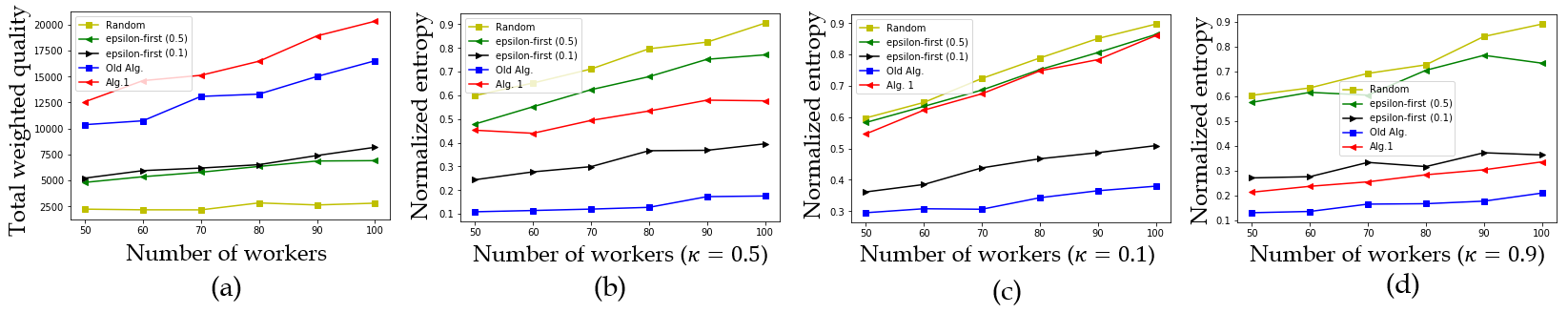}
    \end{center}
    \caption{Simulation results for the effect of the number of workers on the total weighted quality and the normalized entropy.}
    \label{s2}
\end{figure*}

\begin{figure*}[t]
    \begin{center}
        \includegraphics[scale=0.59]{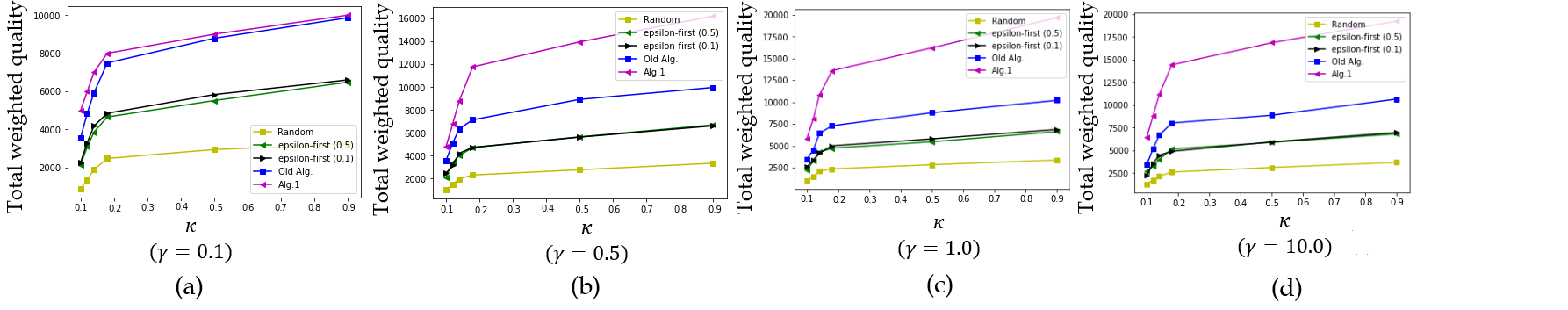}
    \end{center}
    \caption{Simulation results for the total weighted quality with varying our two new parameters $\kappa$ and $\gamma$ together.}
    \label{s4}
\end{figure*}

In addition, we refine the dataset by removing the taxicabs that rarely visit the considered locations. From those $600$ locations, we let $M$ be generated so that $100$ to $600$ locations are selected. Moreover, in our simulations, we select $N$ taxicabs that perform as workers. $N$ is selected to cover the range $[50, 100]$. We set $M=300$, $N=50$, $\kappa=0.4$, $\gamma=1$, $\lambda=5$, $r=2$, $B=850$ $K=\lceil\frac{N}{3}\rceil$ to be the default values used whenever they are not mentioned.

Our simulations were conducted on a machine equipped with an Intel(R) Core(TM) i5-7200U CPU operating at 2.50 GHz and supported by 16.0 GB of installed RAM. This hardware configuration provided a robust platform for evaluating the performance of our algorithm, particularly in terms of iteration time. We observed that the iteration time of the algorithm was significantly influenced by the choice of the accuracy parameter \( r \). Specifically, when \( r \) was set to 2, the algorithm demonstrated a relatively efficient iteration time, aligning with the computational capabilities of the machine. However, increasing \( r \) to 3 resulted in a noticeable increase in the iteration time, reflecting the exponential growth in computational complexity associated with higher \( r \) values. Despite this increase, the machine's processing power was sufficient to handle these computations, albeit with extended execution times. This aspect of our experimental setup shows the importance of considering both algorithmic efficiency and hardware capabilities in practical applications, especially for tasks involving complex computations and large datasets.

Regarding the options chosen for each one of the workers, we construct them by first assigning each worker (i.e. taxicab) with all tasks (i.e. locations) that are within a distance of $200$ meters from them. Afterward, we randomly generate the subsets of options for each one of the workers so that the size of each one of them is randomly generated to be between $5$ and $15$ tasks. Regarding the distribution of the quality values for each worker, we generate them randomly so that their mean values are within the range of $(0,1)$. Regarding the cost value for each one of the options for the workers, we simply set the cost for each option to be linearly proportional with its size. The proportionality factor for each worker is uniformly randomly generated to be in the range $(0,1)$. Lastly, we let the initial weight values for each one of the tasks $w_j^1$ be uniformly-randomly generated so that $\Sigma_{\forall j}w_j^1=1$.

\subsection{Algorithm Comparison}

In this study, we run the new diversity-based overlap-aware model and compare it to the basic UCB-based algorithm, referred to as the Old Algorithm, previously presented in details in \cite{gao}, which solves the problem without consideration to the diversity of the selected workers and without the awareness of the overlaps. Additionally, we consider a basic greedy-$\epsilon$ algorithm, which operates such that a portion of the budget, specifically the first $\epsilon B$, is utilized for exploration through the random selection of workers and options. The remaining budget is then dedicated to exploitation by sticking with the top $K$ options. To perform this comparison, we adopt both $\epsilon=0.1$ and $\epsilon=0.5$. It is worth noting that the value of $\epsilon$ determines the proportion of the budget allocated to exploration versus exploitation. For $\epsilon=0.1$, it is a conservative approach that favors known options (90\% exploitation), ideal for stable environments and quick solution convergence. Conversely, $\epsilon=0.5$ adopts a balanced strategy, equally dividing between exploration and exploitation, suitable for dynamic environments like crowdsensing. This contrast in $\epsilon$ values enables comprehensive algorithm evaluation under diverse conditions, highlighting adaptability in unpredictable settings.

\begin{figure*}[t]
    \begin{center}
        \includegraphics[scale=0.59]{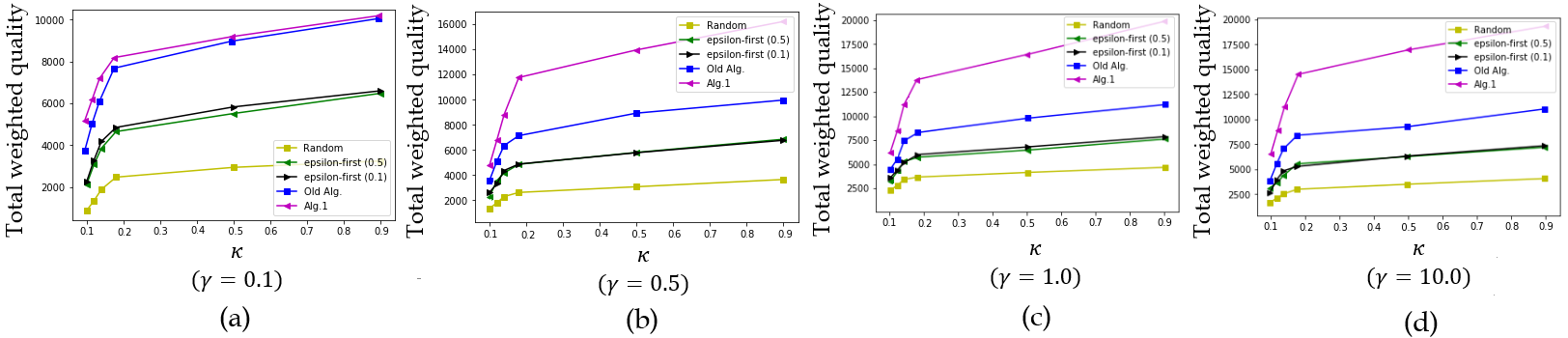}
    \end{center}
    \caption{Simulation results for the total weighted quality with varying our two new parameters $\kappa$ and $\gamma$ together. The value of $r=3.$}
    \label{s5}
\end{figure*}

Regarding the quantification of our novel contribution, the diversity is measured here in terms of the \emph{Normalized Entropy}, which is defined as in Equations \ref{entr1} - \ref{entr2}:

\begin{equation}
    \text{Normalized Entropy}=-\Sigma_{\forall j} (p_j \times \log_{M} p_j),
    \label{entr1}
\end{equation}

\vspace{-3mm}
\begin{equation}
    p_j=m_j^{\tau(B)} /\Sigma_{\forall j}m_j^{\tau(B)}.
    \label{entr2}
\end{equation}

The normalized entropy is an excellent measure of diversity as it ranges in $(0,1]$. Equation \ref{eq1a} provides the formal definition of $m_j^{\tau(B)}$. It is important to note that this measure allows for the quantification of diversity within a given system or dataset.

\subsection{Experimental Results}
As shown in Figure \ref{s1} (b), our proposed algorithm consistently outperforms the old algorithm by approximately 21\% on average for various budget values. This trend is also evident when examining the performance of the algorithms for different values of the percentage of total workers $K$, as depicted in Figure \ref{s1} (c). On average, our algorithm outperforms the old algorithm by around 37\%. It is worth noting that the diversity-based nature of our algorithm leads to a larger number of rounds being required, as shown in plot (d) of Figure \ref{s1}. This is due to the fact that our algorithm takes into account the decline in the weight values of tasks that are repeated.

The impact of the overlapping factor $\gamma$ on the total weighted quality is demonstrated in Figure \ref{s3}. Higher $\gamma$ values increase the value of the completion quality of the tasks in overlaps by making it closer to the summation of the completion qualities of those tasks by all workers and farther from the maximum completion qualities of the tasks in overlaps. With larger values of $\gamma$, the return of overlaps tends to be significantly higher, as is the case in our used traces, which feature a high number of overlaps. This is why we can see that $\gamma$ has a dramatic effect on the total weighted quality. Furthermore, as seen in the figure, the impact of changing the value of $r$ is not as pronounced when $r$ is set to values larger then $2.$ Therefore, it would be more efficient to avoid using high values of $r,$ as increasing this factor significantly increases the time complexity of the algorithm.

One of the most noteworthy results of our analysis can be seen in the plots presented in Figure \ref{s2}, which illustrate the normalized entropy values of the selected tasks over all rounds using different algorithms. It is clear from these plots that the purely random algorithm outperforms all other algorithms in terms of diversity, as it consistently chooses a different random set of $K$ options in each round. The $\epsilon$-greedy algorithm with a value of $\epsilon=0.5$ comes in second place in terms of diversity, as half of the budget is spent on exploring new options by randomly selecting tasks.

Our own proposed algorithm achieves a more diverse set of tasks for lower $\kappa$ values, as this causes the weight values of the tasks to decline more quickly. The old algorithm, on the other hand, performs poorly in terms of diversity as it does not take into account the fact that the weight of a task decreases with repetition. However, for higher values of $\kappa$, the old algorithm performs similarly to our proposed algorithm. Figure \ref{s2} also shows how the total weighted quality of the tasks varies with the number of workers considered.

The next set of plots, presented in Figure \ref{s4}, exhibit the influence of both $\kappa$ and $\gamma$ on the total weighted quality values. It can be seen that higher values of $\kappa$ result in a less severe decline in the weight of the tasks, which allows our proposed algorithm to perform similarly to the old algorithm when $\gamma$ has a relatively low value. Overall, the results of our simulations align with our theoretical predictions and provide valuable insights into the relationships between the various parameters and the performance of the algorithms.

Following the analysis with \( r = 2 \), Figure \ref{s5} presents a new set of plots that depict the algorithm's performance with the accuracy parameter set to \( r = 3 \). Remarkably, the results shown in these plots are almost identical to those observed in Figure \ref{s4}, where \( r = 2 \). This similarity in outcomes highlights the robustness of our algorithm against changes in the accuracy parameter, particularly in terms of managing task overlaps. Despite the higher computational demand associated with \( r = 3 \), the algorithm consistently achieves similar utility values, indicating that increases in \( r \) do not significantly affect the overall performance. This observation also reinforces the efficiency of our approach in selecting optimal task-worker combinations, effectively minimizing unnecessary overlaps, and thus maintaining performance even as the complexity of the problem increases. These findings further substantiate the practical applicability of our algorithm in real-world scenarios, where computational resources and efficiency are crucial considerations.

In addressing the scalability of our algorithm, it is essential to consider the constraints imposed by the CRAWDAD dataset used in our simulations. This dataset, consisting of trace data from approximately 320 taxi cars in Rome, Italy, naturally limits the range of scalability that can be explored. We selected parameters, such as the 200-meter distance to each worker and the number of tasks per worker set between 5 and 15, to encompass the broadest feasible spectrum for this dataset. While these constraints restrict the full extent of scalability testing, our approach was to utilize the dataset to its maximum potential, ensuring a thorough evaluation of our algorithm's performance. Our analysis reveals that the algorithm demonstrates consistent and effective scalability within the dataset's scope, maintaining robust performance as the number of workers and tasks increases. These findings show the adaptability and robustness of our algorithm, suggesting its capability to handle diverse system sizes and conditions within the realistic confines of the available data.

\section{Conclusion}

Mobile crowdsensing attracted researchers in recent years. In this paper, we considered the problem where the sensing qualities of every worker are not known so they need to be learned during the process. Tasks can get covered by more than one worker in the same round. We introduced a novel model in which the completion quality of tasks covered by more than one worker can range from the maximum individual sensing quality to the total of all sensing qualities of the workers covering it. In addition, we modeled the importance value for each task to decline as the task is sensed more over the rounds in order to encourage a more diverse set of tasks to be covered. We designed a budgeted diversity-driven overlap-aware strategy to recruit the workers in a way that maximizes the total weighted qualities encouraging the diversity of tasks covered over rounds with the consideration of overlaps. We used a combinatorial multi-armed bandit setting to model this problem. We utilized an extended version of the upper confidence bound to develop the recruitment strategy and showed the regret analysis. Finally, we showed a comprehensive simulation applied to real data that showed how our algorithm excels compared to other algorithms. The simulation showed that our algorithm gives 21\% more total weighted quality than the existing ones considering this model for different budget values.

\vskip 2mm
\zihao{5}
\noindent
\textbf{Acknowledgment}
\vskip 2mm

\zihao{5--}
\noindent
This research was supported in part by NSF grants CPS 2128378, CNS 2107014, CNS 2150152, CNS 1824440, CNS 1828363, and CNS 1757533.

\vskip 2mm
\zihao{5}
\noindent

\renewcommand\refname{\zihao{5}

\noindent\parbox{8.3cm}{\parpic{\includegraphics[scale=0.4]{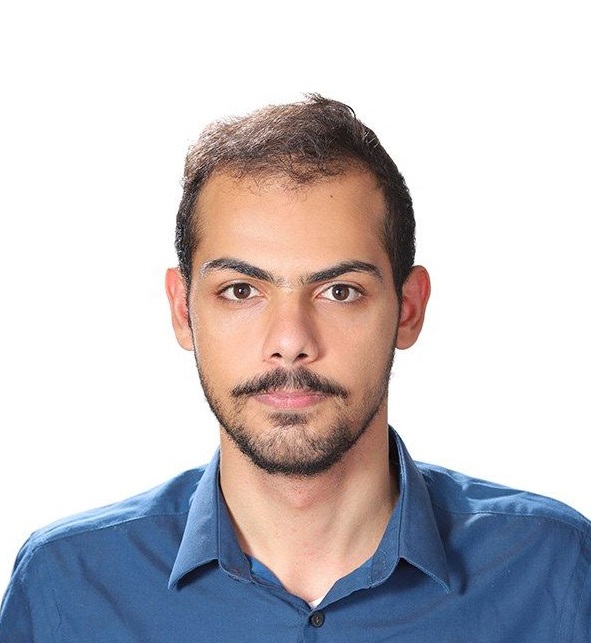}}{\small\quad {\bf Abdalaziz Sawwan} is a Ph.D. student in Computer and Information Sciences at Temple University. Sawwan received his bachelor's degree in Electrical Engineering from the University of Jordan in 2020. His current research interests include learning theory, multi-armed bandits, crowdsensing, mobile charging and wireless networks, and routing protocols and Age of Information.}\\[1mm]}

\noindent\parbox{8.3cm}{\parpic{\includegraphics[scale=0.35]{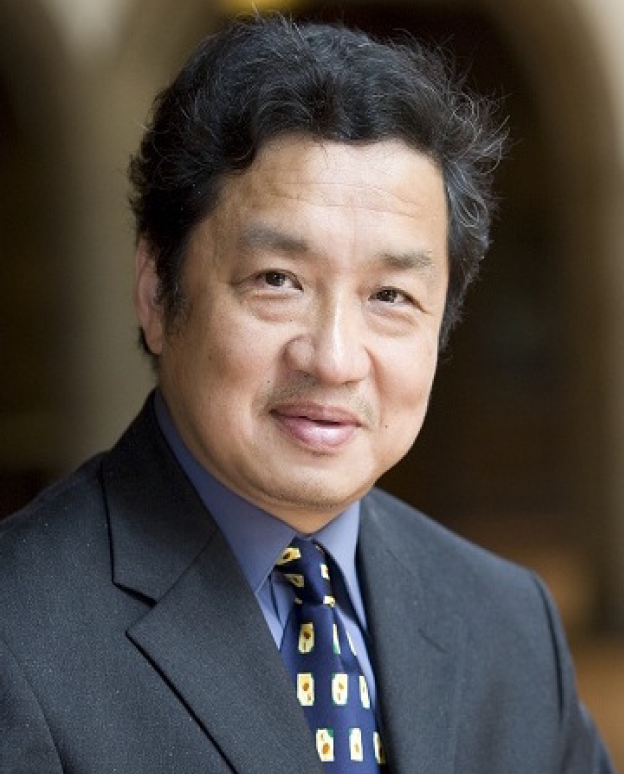}}{\small\quad {\bf Jie Wu} is the director of the Center for Networked Computing and Laura H. Carnell professor at Temple University. He also serves as the director of International Affairs at College of Science and Technology. He served as chair of Department of Computer and Information Sciences from the summer of 2009 to the summer of 2016 and associate vice provost for International Affairs from the fall of 2015 to the summer of 2017. Prior to joining Temple University, he was a program director at the National Science Foundation and was a distinguished professor at Florida Atlantic University. His current research interests include mobile computing and wireless networks, routing protocols, cloud and green computing, network trust and security, and social network applications. Dr. Wu regularly publishes in scholarly journals, conference proceedings, and books. He serves on several editorial boards, including IEEE Transactions on Mobile Computing, IEEE Transactions on Service Computing, Journal of Parallel and Distributed Computing, and Journal of Computer Science and Technology. Dr. Wu was general co-chair for IEEE MASS 2006, IEEE IPDPS 2008, IEEE ICDCS 2013, ACM MobiHoc 2014, ICPP 2016, and IEEE CNS 2016, as well as program cochair for IEEE INFOCOM 2011 and CCF CNCC 2013. He was an IEEE Computer Society Distinguished Visitor, ACM Distinguished Speaker, and chair for the IEEE Technical Committee on Distributed Processing (TCDP). Dr. Wu is a fellow of the AAAS and a fellow of the IEEE. He is the recipient of the 2011 China Computer Federation (CCF) Overseas Outstanding Achievement Award. }\\[1mm]}


\mbox{}

\end{document}